%% file: main.tex
\documentclass[10pt,twocolumn,letterpaper]{article}

\usepackage{cvpr}              % To produce the CAMERA-READY version
\usepackage{amsmath,amsfonts}
\usepackage{algorithm}
\usepackage{algorithmic}
\usepackage{multirow}
\usepackage{colortbl}
\usepackage{arydshln} % for dashed line
\usepackage{booktabs}
\usepackage{subcaption}
\usepackage{nicematrix} % part column color
\usepackage{tablefootnote}

\usepackage{array}
\usepackage{textcomp}
\usepackage{stfloats}
\usepackage{url}
\usepackage{verbatim}
\usepackage{graphicx}
\usepackage{cite}
\usepackage{tabularx}
\usepackage{xcolor}
\usepackage[algo2e,linesnumbered,noend]{algorithm2e}
\usepackage{graphicx}
\usepackage{amssymb}
\usepackage{epstopdf}
\usepackage{multicol}
\usepackage{tabularx}
\usepackage{caption}
\usepackage{enumitem}
\usepackage{booktabs}
\usepackage{pifont}
\usepackage{bm}
\usepackage{wrapfig}
\usepackage{amsmath}
\usepackage{pifont}
\usepackage{lipsum}
\usepackage{dsfont}
\usepackage{mathrsfs}
\usepackage{amsthm}
\usepackage{float}
\usepackage{microtype}
\usepackage{graphicx}
\usepackage{array} % 必需
\usepackage{makecell}
\usepackage{booktabs} % for professional tables
\usepackage{wrapfig}
\usepackage{blindtext}
\usepackage{enumitem}
\usepackage{adjustbox}
\usepackage{booktabs}
\usepackage[table]{xcolor}% http://ctan.org/pkg/xcolor
\usepackage{algorithm}
\definecolor{aliceblue}{RGB}{176,223,229}

\definecolor{zzblue}{RGB}{156,183,245}

\usepackage{subcaption}
\usepackage{natbib}
\usepackage[accsupp]{axessibility}
\usepackage{amsmath}
\usepackage{amssymb}
\usepackage{mathtools}
\usepackage{amsthm}
\usepackage{wrapfig}
\theoremstyle{plain}

\theoremstyle{definition}

\theoremstyle{remark}

\definecolor{C0}{rgb}{0.121569, 0.466667, 0.705882}
\definecolor{C1}{rgb}{1.000000, 0.498039, 0.054902}
\definecolor{C2}{rgb}{0.172549, 0.627451, 0.172549}
\definecolor{C3}{rgb}{0.839216, 0.152941, 0.156863}
\definecolor{C4}{rgb}{0.580392, 0.403922, 0.741176}
\definecolor{C5}{rgb}{0.549020, 0.337255, 0.294118}
\definecolor{C6}{rgb}{0.890196, 0.466667, 0.760784}
\definecolor{C7}{rgb}{0.498039, 0.498039, 0.498039}
\definecolor{C8}{rgb}{0.737255, 0.741176, 0.133333}
\definecolor{C9}{rgb}{0.090196, 0.745098, 0.811765}
\newcommand\ve[1]{\mathbf{#1}}
\definecolor{iccvblue}{rgb}{0.21,0.49,0.74}

\DeclareMathOperator*{\argmin}{arg\,min}

\makeatletter
\renewcommand{\@cite}[2]{\textcolor{blue}{[#1\if@tempswa , #2\fi]}}
\makeatother
\usepackage[T1]{fontenc}
\newlength\savewidth

\renewcommand{\paragraph}[1]{\vspace{1.25mm}\noindent\textbf{#1}}

\newcolumntype{x}[1]{>{\centering\arraybackslash}p{#1pt}}
\newcolumntype{y}[1]{>{\raggedright\arraybackslash}p{#1pt}}
\newcolumntype{z}[1]{>{\raggedleft\arraybackslash}p{#1pt}}

\newcommand{\app}{\raise.17ex\hbox{$\scriptstyle\sim$}}

\definecolor{deemph}{gray}{0.6}

\definecolor{baselinecolor}{gray}{.9}

\definecolor{zeroshotcolor}{gray}{.3}

\definecolor{LightCyan}{rgb}{0.92,1,1}
\newcommand{\cmark}{\ding{51}}
\newcommand{\xmark}{\ding{55}}
\newcommand{\best}[1]{$\mathbf{#1}$}

\definecolor{demphcolor}{RGB}{144,144,144}

\definecolor{LightCyan}{rgb}{0.92,1,1}
\definecolor{darkergreen}{RGB}{21, 152, 56}
\definecolor{red2}{RGB}{252, 54, 65}

\definecolor{bluebell}{rgb}{0.84, 0.84, 0.92}
\newcommand*\colorcmark[1]{%
  \expandafter\newcommand\csname #1cmark\endcsname{\textcolor{#1}{\ding{51}}}%
}
\colorcmark{green}

\newcommand*\colorxmark[1]{%
  \expandafter\newcommand\csname #1xmark\endcsname{\textcolor{#1}{\ding{55}}}%
}
\definecolor{Highlight}{HTML}{39b54a}  % green

\colorxmark{red}
\definecolor{cvprblue}{rgb}{0.21,0.49,0.74}
\usepackage[pagebackref,breaklinks,colorlinks,allcolors=cvprblue]{hyperref}

\def\paperID{34724} % *** Enter the Paper ID here
\def\confName{CVPR}
\def\confYear{2026}

\title{FastLightGen: Fast and Light Video Generation with Fewer Steps and Parameters}
\author{Shitong Shao$^{1}$ \qquad Yufei Gu$^{1}$ \qquad Zeke Xie$^{1,*}$ \\
$^1$The Hong Kong University of Science and Technology (Guangzhou), Guangzhou, China \\
{\tt\small sshao213@connect.hkust-gz.edu.cn, ygu167.gu@connect.hkust-gz.edu.cn, zekexie@hkust-gz.edu.cn} \\
{\tt\small $*$:Corresponding author}}

\begin{document}

\maketitle
\input{sec/0_abstract}
%Video Generation, Diffusion Models, Step and Size Joint Distillation, Well-judged Teacher Guidance

\input{sec/1_intro}

\input{sec/2_preliminary}

\input{sec/3_method}
\input{sec/4_experiment}
\input{sec/5_conclusion}

{
    \small
    \bibliographystyle{ieeenat_fullname}
    \bibliography{main}
}

\input{sec/appendix}

\end{document}

%% file: sec/0_abstract.tex
\begin{abstract}
The recent advent of powerful video generation models, such as Hunyuan, WanX, Veo3, and Kling, has inaugurated a new era in the field. However, the practical deployment of these models is severely impeded by their substantial computational overhead, which stems from enormous parameter counts and the iterative, multi-step sampling process required during inference. Prior research on accelerating generative models has predominantly followed two distinct trajectories: reducing the number of sampling steps (e.g., LCM, DMD, and MagicDistillation) or compressing the model size for more efficient inference (e.g., ICMD). The potential of simultaneously compressing both to create a fast and lightweight model remains an unexplored avenue. In this paper, we propose \textit{\textbf{FastLightGen}}, an algorithm that transforms large, computationally expensive models into fast, lightweight counterparts. The core idea is to construct an optimal teacher model, one engineered to maximize student performance, within a synergistic framework for distilling both model size and inference steps. Our extensive experiments on HunyuanVideo-ATI2V and WanX-TI2V reveal that a generator using 4-step sampling and 30\% parameter pruning achieves optimal visual quality under a constrained inference budget. Furthermore, FastLightGen consistently outperforms all competing methods, establishing a new state-of-the-art in efficient video generation.
\end{abstract} 
% introduces a strategy we term ``well-guided teacher guidance''. This approach

%% file: sec/1_intro.tex
\section{Introduction}
\label{sec:intro}

\begin{figure}[t]
    \centering
    \includegraphics[width=0.9\linewidth,trim={0 0 0cm 0},clip]{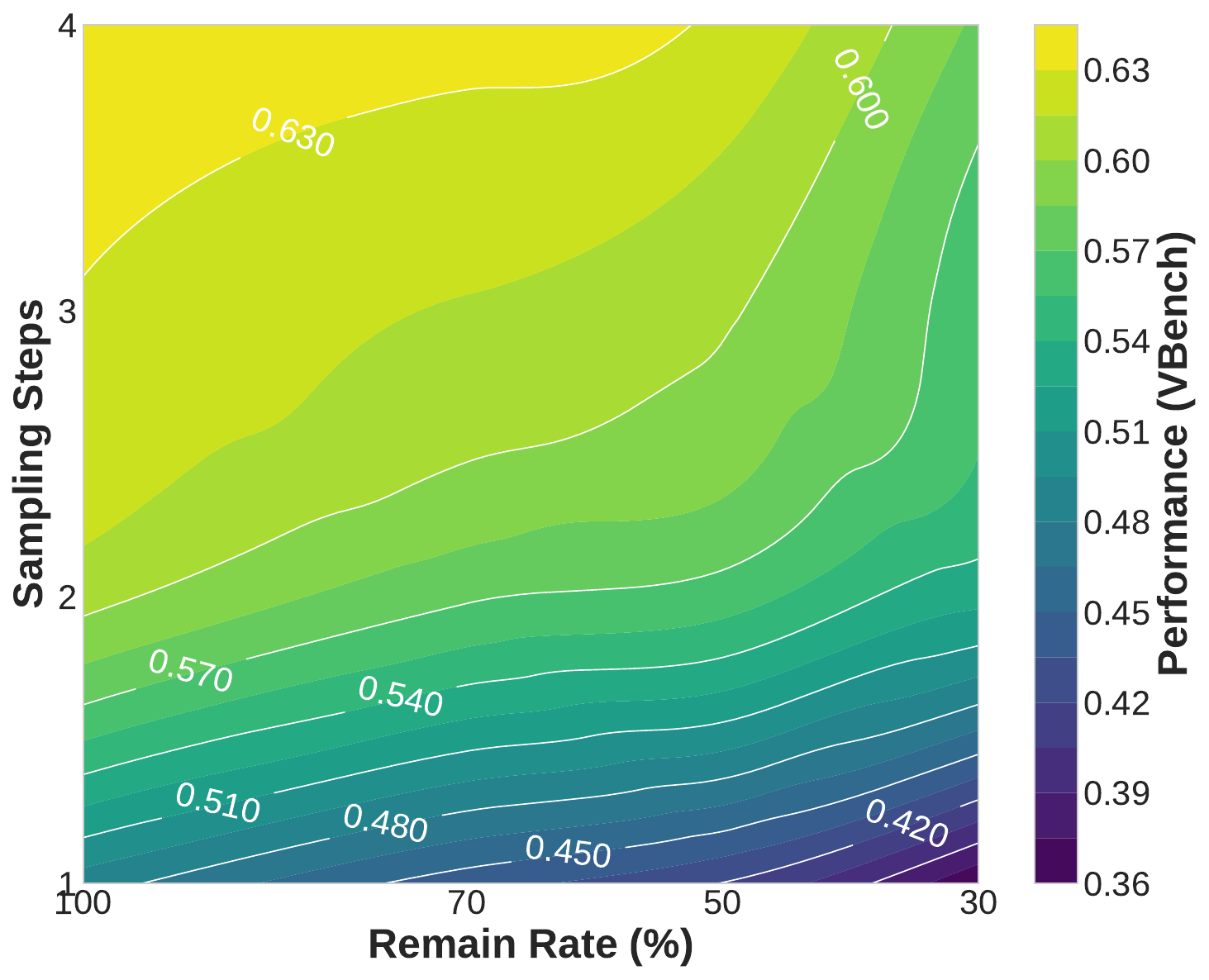}
    \vspace{-15pt}
    \caption{
    The model performance is calculated as the average of Vbench's dynamic degree, aesthetic, and image quality metrics, plotted against sampling steps and parameter retention rate. The parameter-sampling steps trade-offs in FastLightGen compression
    reveal that performance is initially more sensitive to a reduction in sampling steps than to parameter pruning. As the overall compression ratio increases, the performance degradation from either method begins to converge. Notably, the model with only 30\% of its parameters at 4 steps performs on par with the 100\% parameter model at 1.2 steps. We report that a moderate compression strategy, specifically, retaining 70\% of parameters at 4 sampling steps—strikes a highly effective trade-off, achieving a theoretical speedup of $\approx$35.71$\times$ over the 50-step unpruned baseline which relies on classifier-free guidance (CFG)~\citep{nips2021_classifier_free_guidance}.
    }
    \label{fig:distilling_laws}
    \vspace{-15pt}
\end{figure}

The exceptional performance of recent video generation models~\citep{animatediff,cogvideox,chen2023videocrafter1,kong2024hunyuanvideo,wan2025wan,genmo2024mochi,d2c,w2sd,golden_noise,shao2025ivmixed,gao2025principled,li2025diffusion,zigzag,10.1145/3796982} is largely attributed to their large-scale architectures and iterative, multi-step sampling processes. For example, HunyuanVideo~\citep{kong2024hunyuanvideo} and WanX~\citep{wan2025wan} utilize diffusion transformer (DiT)~\citep{DIT} architectures with over 13 billion parameters and employ a multi-step, flow-matching-based denoising process~\citep{ddim,ddpm_begin,sde} to synthesize high-quality videos. This reliance on massive parameter counts and multi-step denoising, while crucial for performance, imposes a substantial computational burden that presents a major barrier to practical deployment. To illustrate, synthesizing a 5-second video with models like HunyuanVideo~\citep{kong2024hunyuanvideo} or WanX~\citep{wan2025wan} requires approximately 20 minutes of inference time on a single NVIDIA H100 GPU. This degree of latency is prohibitively high for most real-world applications, making it untenable for both service providers and end-users~\citep{fastvideo}.

Prior work has identified two primary sources of the substantial computational overhead in video generation models~\citep{wu2024individual,dmd,dmd2,yin2025causvid}: \textbf{(1)} high parameter counts and \textbf{(2)} iterative, multi-step sampling procedures. To address the challenge of high parameter counts, one direct approach is to train a lightweight model from scratch. However, this method typically leads to a significant degradation in visual quality and motion dynamics. For example, while models like WanX-1.3B are parameter-efficient, the quality of their generated videos is often insufficient for practical applications~\citep{wan2025wan}. To mitigate the computational overhead of multi-step sampling, the prevailing approach is step distillation~\citep{iclr22_progressive,icml23_consistency,iclr22_rect}. In the context of video generation, few-step generators derived from the distribution matching distillation (DMD)~\citep{dmd,dmd2} framework have demonstrated particularly effective. Recent studies, such as FastVideo~\citep{fastvideo} and MagicDistillation~\citep{shao2025magicdistillation}, have built upon this foundation. While these methods can synthesize high-quality images in few-step scenarios (e.g., 4-step sampling), their performance degrades significantly when the step count is reduced to just one or two.

\begin{figure}[t]
    \centering
    \includegraphics[width=0.9\linewidth,height=6cm,trim={5 0 0cm 0},clip]{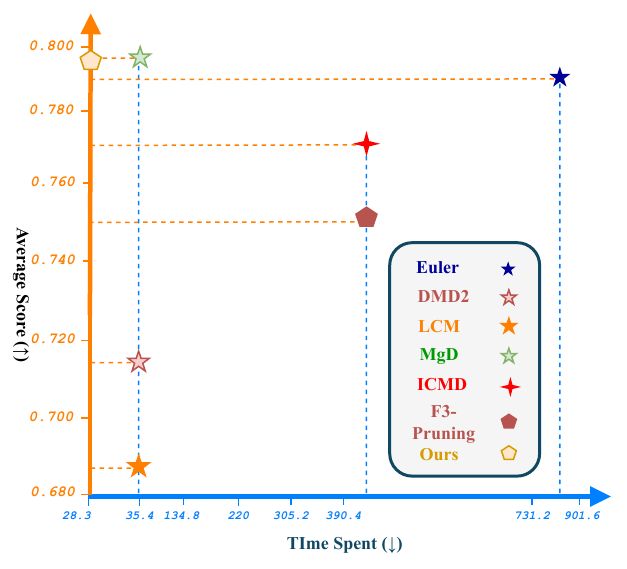}
    \vspace{-15pt}
    \caption{Vbench average score vs. time spent. MgD: MagicDistillation. Among the accelerated sampling algorithms evaluated, our proposed FastLightGen achieves the greatest speedup, while its performance also surpasses that of the teacher model (Euler).}
    \label{fig:easy_framework}
    \vspace{-10pt}
\end{figure}

Prior video diffusion model acceleration methods have typically addressed sampling step distillation and model compression in isolation, largely overlooking their joint optimization. We argue that there is significant potential in a co-distillation approach that jointly tunes the number of sampling steps and the student model's size to yield maximal performance gains for a given computational cost. The visualization results in Fig.~\ref{fig:distilling_laws} clearly demonstrate that compared to performing compression at a single dimension (i.e., size or step), joint distillation enables student models to become faster and more lightweight while achieving the same performance gains. To be specific, the advantages of co-distillation are clear when targeting a specific performance threshold, e.g., 0.630. To reach this target, step distillation alone produces a 3-step unpruned model, achieving a 33.3$\times$ speedup. In contrast, the joint distillation yields a 4-step model retaining only 50\% of the parameters, achieving a substantially higher 50$\times$ speedup. Therefore, developing a method that jointly distills sampling steps and model size is a crucial next step to address this research gap.

\begin{figure*}[t]
    \centering
    \includegraphics[width=0.9\linewidth]{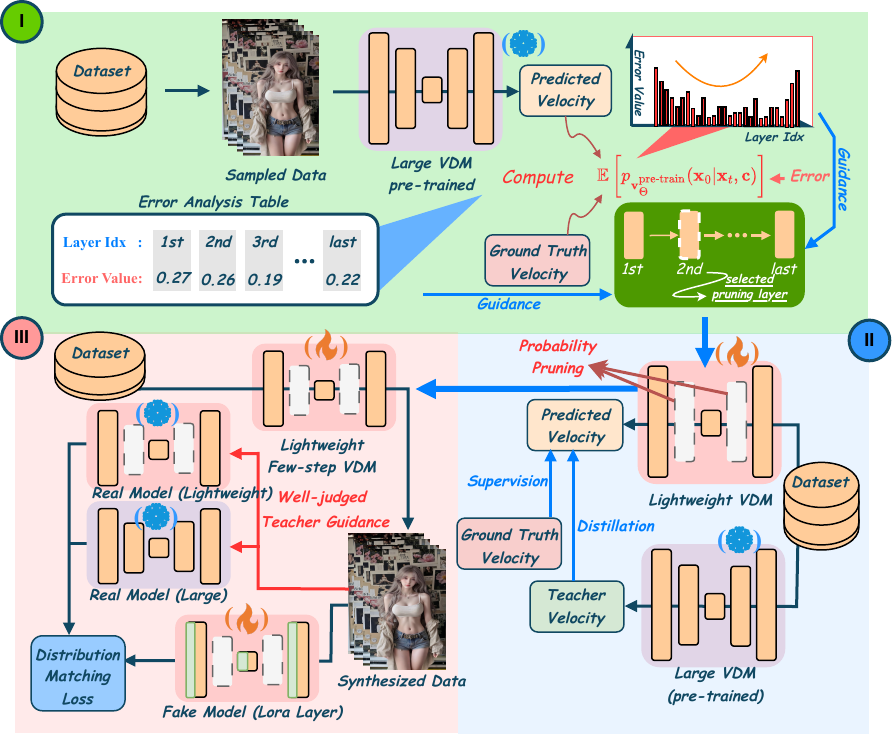}
    \vspace{-10pt}
    \caption{Overview of three-stage distillation pipeline \textit{\textbf{FastLightGen}}. FastLightGen begins by identifying less critical layers within the model. The second stage introduces dynamic probabilistic pruning, where these identified layers are stochastically skipped during training. This process yields a robust, stochastically-pruned student model for the final stage. In this final stage, we perform distribution matching. Our "well-guided teacher guidance", which is constructed from the stochastically-pruned student, ensures that the resulting lightweight, few-step generator maintains high performance.}
    \label{fig:total_framework_lightning}
    \vspace{-10pt}
\end{figure*}

Based on these considerations, we propose \textbf{\textit{FastLightGen}}, a three-stage algorithm that operates by first constructing a compact model and then distilling knowledge into it. The algorithm proceeds in three stages: \textbf{(1)} identifying unimportant model blocks, \textbf{(2)} training a robust, dynamically pruned model, and \textbf{(3)} performing a fine-grained distribution matching using a novel guidance mechanism. In the initial stage, we identify the least important model blocks by systematically skipping each one and leveraging Tweedie's formula to estimate the evidence lower bound (ELBO). Blocks causing the smallest ELBO drop are marked as redundant. The second stage employs dynamic probabilistic pruning, where the identified unimportant blocks are stochastically dropped during training. This process yields a single, robust model that adapts to a dynamic architecture while maintaining high sample quality. In the final stage, the model from the preceding phase is used to initialize four components: a student (a few-step generator), and a teacher apparatus composed of a strong teacher, a weak teacher, and a trainable fake DiT. The three teacher components are parameter-sharing, while the student model is separate. As illustrated in Fig.~\ref{fig:total_framework_lightning}, the strong and weak teachers are two configurations derived from the same underlying model.

% In comparative experiments, FastLightGen significantly surpasses prior methods like DMD2~\citep{dmd,dmd2}, MagicDistillation~\citep{shao2025magicdistillation}, and ICMD~\citep{wu2024individual} in both generation quality and speedup. Remarkably, on the VBench-I2V benchmark, FastLightGen even outperforms its own teacher model, WanX-TI2V.

\noindent{Our contribution consists of the following three parts:}
\begin{enumerate}[nosep]
    \item We demonstrate for the first time that a synergistic distillation of model size and inference steps offers significant advantages over distilling either modality in isolation (Fig.~\ref{fig:distilling_laws}). This result motivates a re-examination of accelerated sampling techniques for visual generative models.
    \item We propose FastLightGen, a novel three-stage VDM distillation pipeline that distills large teacher models into compact, efficient student models. Concurrently, we introduce a novel teacher guidance strategy to configure the most suitable teacher model for the student, thereby maximizing the student's final performance.
    \item As illustrated in Fig.~\ref{fig:easy_framework}, extensive comparative experiments show that FastLightGen outperforms nearly all existing accelerated sampling algorithms and even surpasses its own teacher model, achieving the best performance.
\end{enumerate}

%% file: sec/2_preliminary.tex
\section{Related Work}
\label{sec:preliminaries}

\paragraph{Large-Scale Open-Source VDMs.} Recent state-of-the-art open-source VDMs have increasingly converged on a common architectural blueprint, exemplified by leading works such as WanX~\citep{wan2025wan}, Step-Video~\citep{stepfunvideo}, and HunyuanVideo~\citep{kong2024hunyuanvideo}. A notable trend is the adoption of the DiT backbone over traditional UNet architectures, which offers superior scaling properties and a higher performance ceiling. For the generative process, these models leverage the flow-matching paradigm~\citep{flow_matching,iclr22_rect,xie2026guidance} to define a continuous-time mapping from a Gaussian noise distribution $p_1(\ve{x}_1)$ to the real data distribution $p_0(\ve{x}_0)$. This mapping is governed by an ordinary differential equation (ODE), $d\ve{x}_t = \ve{v}_\Theta(\ve{x}_t,t) dt$, where the velocity field $\ve{v}_\Theta$ is trained by minimizing a least squares objective:
\begin{equation}
    \begin{split}
    & \mathop{\arg\min}_{\Theta} \int_0^1 \mathbb{E}\left[\alpha_t\Vert \partial p_t(\ve{x}_t)/\partial t - \ve{v}_\Theta(\ve{x}_t,t)\Vert_2^2\right]dt, \\
    \end{split}
    \label{eq:hunyuan2}
\end{equation}
with $\alpha_t$ as a weighting factor~\citep{karras2022elucidating,ddpm_begin}. After training, video synthesis is performed directly via the Euler algorithm. Furthermore, these models employ a consistent VAE design, compressing video inputs from a resolution of $T\!\times\!H\!\times\!W$ to a latent representation of size $\lfloor\!\frac{T}{4}\!\rfloor\!\times\!\lfloor\frac{H}{8}\!\rfloor\!\times\!\lfloor\!\frac{W}{8}\!\rfloor$.

\paragraph{Step Distillation.} Step distillation techniques aim to accelerate diffusion model inference by reducing the required number of function evaluations (NFEs). For video generation, current research in this area is largely dominated by two distinct paradigms. The first paradigm involves aligning the outputs of a few-step student model with those of a multi-step teacher. This approach is best exemplified by latent consistency models (LCMs), which have become a foundational training framework for many open-source, few-step video diffusion models (VDMs), including MCM~\citep{mcm_accelerate}, the T2V-turbo series~\citep{t2vturbov2}, and FastVideo~\citep{fastvideo}. A second, alternative paradigm is distribution matching, which directly optimizes the student to replicate the teacher's output distribution. This strategy has demonstrated a performance advantage over consistency-based alignment in the image synthesis domain, as evidenced by recent methods like DMD~\citep{dmd}, DMD2~\citep{dmd2}, and SiD-LSG~\citep{sid_lsg}. In particular, DMD2 has emerged as a leading step distillation algorithm for video generation due to its clear advantages over prior methods. Specifically, it offers superior performance compared to LCM while being more memory-efficient than SiD. Its effectiveness has established it as a foundational approach, inspiring subsequent video distribution matching algorithms such as MagicDistillation~\citep{shao2025magicdistillation} and rCM~\citep{zheng2025large}.

\paragraph{Size Distillation.} Size distillation for diffusion models typically relies on pruning techniques~\citep{XiaXC22_pruning,ZhuoWLW022_pruning,LagunasCSR21_pruning,zhao2026late}. These methods are broadly classified by their granularity into unstructured~\citep{DongCP17_unstr_pruning,Sanh0R20_unstr_pruning,ParkLMS20_unstr_pruning} and structured~\citep{DingDGH19_str_pruning,YouYYM019_str_pruning,LiuZKZXWCYLZ21_str_pruning} approaches. While unstructured pruning provides fine-grained control, its resulting irregular sparsity patterns are difficult to accelerate on modern hardware. Structured pruning, conversely, operates at a coarser, block-level granularity, thereby yielding practical, hardware-friendly inference speedups. Despite being a common technique for image generation, pruning remains largely underexplored in the video domain. To our knowledge, only two methods, F3-Pruning~\citep{su2024f3} and ICMD~\citep{wu2024individual}, have previously explored this direction. Among them, F3-Pruning prunes weights based on a thresholded ranking of aggregated temporal attention values. ICMD, in contrast, introduces intermediate-layer alignment between teacher and student models and simultaneously applies adversarial training to enhance the motion dynamics of the generated videos.

\begin{figure*}[t]
    \centering
    \includegraphics[width=1\linewidth]{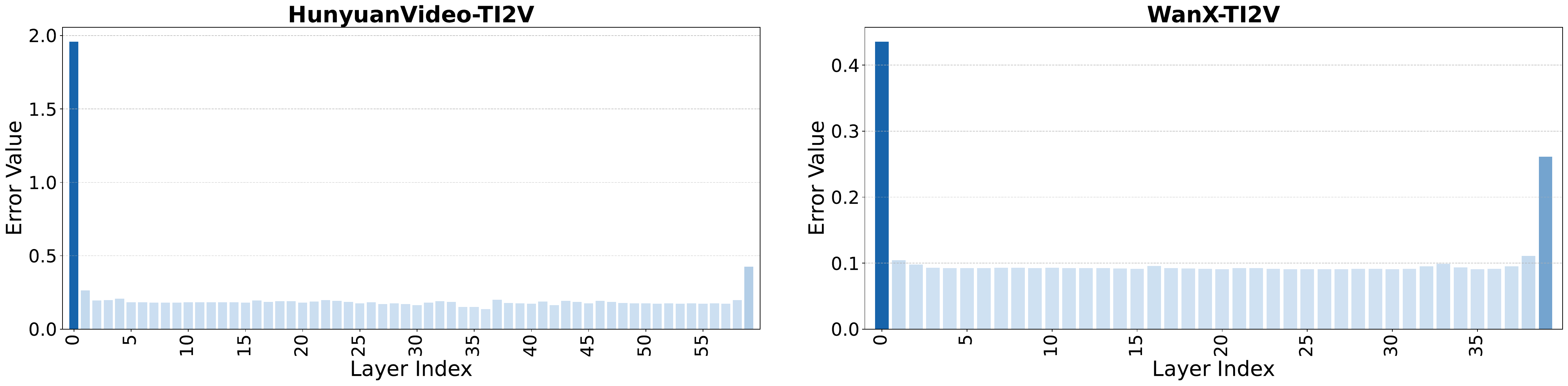}
    \vspace{-20pt}
    \caption{In the first phase of FastLightGen, we identify non-critical layers by visualizing their importance, where a higher error value corresponds to greater layer criticality. For both HunyuanVideo-TI2V and WanX-TI2V, the results reveal a consistent pattern: the initial and final layers are the most critical, while the importance of the intermediate layers is substantially lower.}
    \label{fig:critical_analysis}
    \vspace{-10pt}
\end{figure*}

%% file: sec/3_method.tex
\section{\textbf{FastLightGen} Framework}
\label{sec:method}

\subsection{Overview}

In this section, we introduce FastLightGen, our three-stage distillation algorithm, illustrated in Fig.~\ref{fig:total_framework_lightning}. In the first stage, we identify unimportant layers within the teacher model and analyze their distribution patterns. Next, in the second stage, we employ dynamic probabilistic pruning to train a robust model capable of bypassing these non-essential layers. Finally, in the third stage, we perform a joint step-and-size distillation via distribution matching.

\begin{figure*}[t!]
    \centering
    \includegraphics[width=1\linewidth]{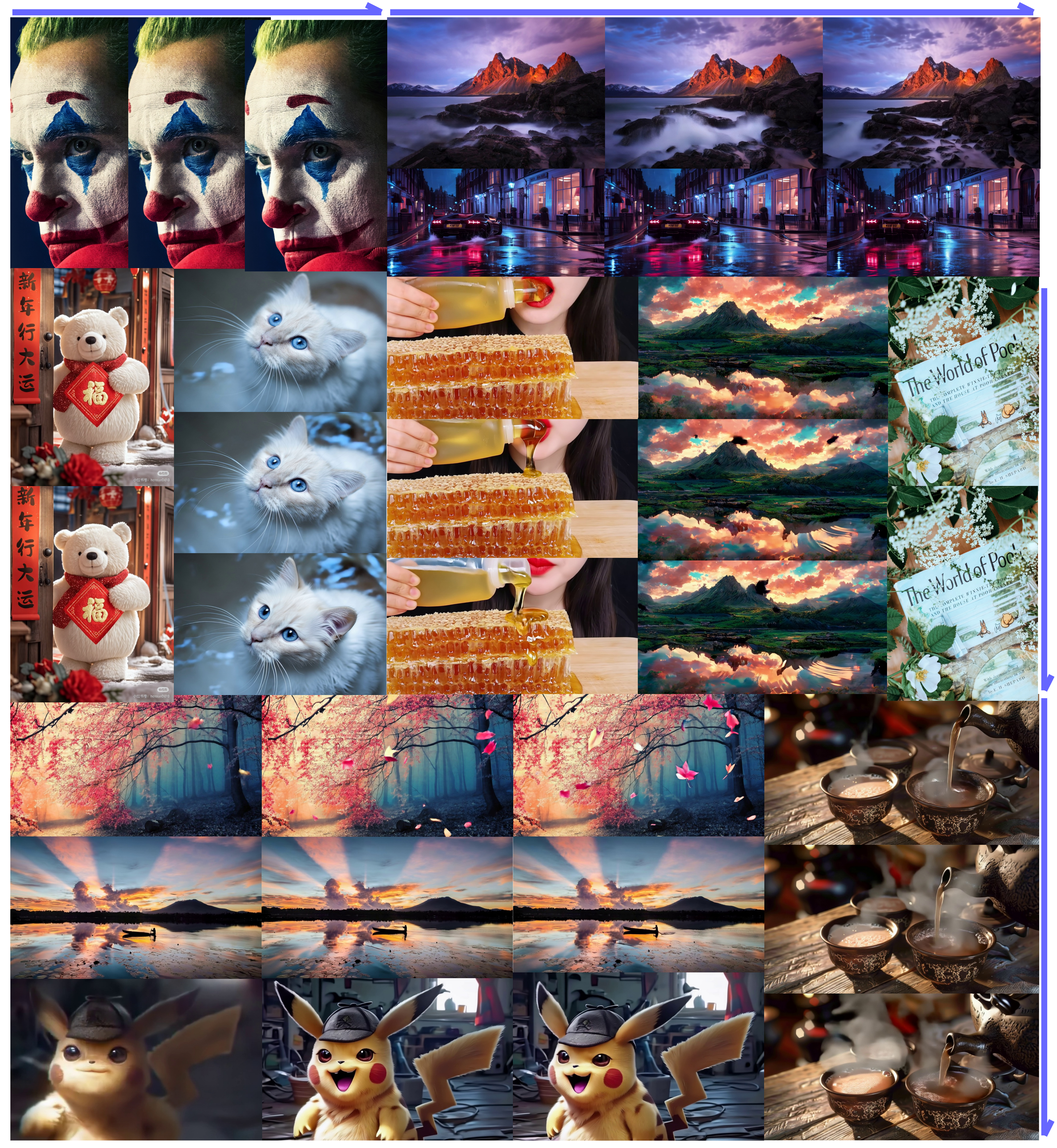}
    \vspace{-20pt}
    \caption{Visualization of FastLightGen (i.e., 4-step generator that retains 70\% of the parameters) across diverse scenarios, including landscapes, food vlogging, dance, and daily activities. Our model generates high-fidelity videos characterized by realistic character motion, detailed expressions, and strong temporal dynamics.}
    \label{fig:visualization_best_performance}
    \vspace{-10pt}
\end{figure*}

\subsection{Stage I: Identifying Unimportant Model Blocks}
Identifying non-essential layers is a critical first step in compressing large-scale video generation models. We consider a DiT model, $\ve{v}_\Theta(\cdot,\cdot)$, composed of N sequential DiT blocks $\{\mathcal{B}^i(\cdot)\}_{i}$. Its forward pass can be formulated as
\begin{equation}
\begin{split}
    y^\text{original} =  g_2\circ \mathcal{B}^{N-1} \circ \cdots \circ \mathcal{B}^2 \circ \mathcal{B}^1 \circ \mathcal{B}^0 \circ g_1(x),
\end{split}
\end{equation}
where $\circ$ denotes the composition, $g_1$ and $g_2$ are auxiliary operations whose computational overheads are negligible compared to those of the DiT blocks. We seek to find an optimal list of blocks $M$, of size $N_\text{short}<N$, by minimizing the following objective:
\begin{equation}
\begin{split}
    &y^\text{pruned} = g_2\circ \mathcal{B}^{M_{N_\text{short}-1}} \circ \cdots \circ \mathcal{B}^{M_1} \circ \mathcal{B}^{M_0} \circ g_1(x), \\
    &\argmin_{M} \mathbb{E}_{x}[\Vert y^\text{original} - y^\text{pruned}\Vert_2^2].
\end{split}
\end{equation}This is an NP-hard problem, yet we can employ a greedy algorithm to solve it. Specifically, for each block $\mathcal{B}_i$, we skip that block and compute its loss $\mathcal{E}^i$ at $x_0$ using Tweedie's formula, which is an estimation of the ELBO~\citep{sde,ddpm_begin}:
\begin{equation}
\begin{split}
    & \mathcal{E}^i = \mathbb{E}_{(t,x_t,x_0)}\Vert x_t - t\ve{v}_\Theta^{\text{skip-}i}(x_t,t) -x_0\Vert_2^2,\\
\end{split}
\end{equation}
where $\ve{v}_\Theta^{\text{skip-}i}(\cdot,\cdot)$ is the DiT with its $i$-th block skipped, and $x_0$ denotes the ground truth data. After computing the loss value for each block, we sort them in descending order to create a ranked sequence $S$. The magnitude of the loss serves as an importance score for the corresponding layer. Consequently, to retain $N_\text{short}$ blocks, the set $M$ is composed of the indices corresponding to the top $N_\text{short}$ entries in $S$.

Our analysis reveals a distinct U-shaped pattern of block importance in VDMs. As illustrated in Fig.~\ref{fig:critical_analysis}, for both HunyuanVideo-TI2V and WanX-TI2V, the initial and final layers are consistently the most critical, with the initial layers showing marginally higher importance scores. This pattern suggests that the intermediate layers contribute less to the final output, potentially under-utilizing the full generative capacity of their DiT blocks. Additionally, HunyuanVideo-TI2V employs a hybrid architecture, consisting of 20 double DiT blocks followed by 40 single DiT blocks. The double block utilizes a multi-modal DiT architecture, while the single block employs the classic DiT architecture. As shown in Fig.~\ref{fig:critical_analysis}, the double blocks yield a higher average loss value, indicating they are more effective at capturing critical information than their single-block counterparts.

\subsection{Stage II: Training a Robust, Dynamically Pruned Model}

Our initial exploration showed that merely skipping non-critical layers for inference results in a failure to generate plausible videos. We argue that the key is to fine-tune this pruned architecture, enabling it to execute the full multi-step diffusion sampling process effectively, relying solely on its remaining critical layers. Following Stage I, we define two models: the pruned model ($\ve{v}_\Theta^\text{pruned}$), which skips unimportant layers, and the unpruned model ($\ve{v}_\Theta^\text{unpruned}$), which retains all layers. During training, non-essential layers are dynamically skipped according to a Bernoulli distribution with $p=0.5$. This process constructs the parameter-sharing unpruned model, $\ve{v}_\Theta^\text{unpruned}$, and the pruned sub-model, $\ve{v}_\Theta^\text{pruned}$. Our objective is to train on real data such that both models can generate high-fidelity videos, while ensuring that $\ve{v}_\Theta^\text{pruned}$ exhibits no perceptible visual degradation compared to $\ve{v}_\Theta^\text{unpruned}$:
\begin{equation}
    \begin{split}
    \mathcal{L}^\text{stage-II}_\text{training} =& \mathbb{E}_{t,x_0,x_1\sim\mathcal{N}(0,\mathbf{I})}\\
    & [(1-\alpha)\Vert\ve{v}_\Theta^\text{pruned}(x_t,t) - (x_0 - x_1)\Vert_2^2 \\
        & + (1-\alpha)\Vert\ve{v}_\Theta^\text{unpruned}(x_t,t) - (x_0 - x_1)\Vert_2^2],\\
    \end{split}
\end{equation}
where $(1-\alpha)$ refers to the training weight. Additionally, we align the output of the pruned model $\ve{v}_\Theta^\text{pruned}$ with that of its unpruned counterpart $\ve{v}_\Theta^\text{unpruned}$:
\begin{equation}
    \begin{split}
    \mathcal{L}^\text{stage-II}_\text{distill} =& \mathbb{E}_{t,x_0,x_1\sim\mathcal{N}(0,\mathbf{I})}\\ &[\alpha\Vert\ve{v}_\Theta^\text{pruned}(x_t,t) - \texttt{SG}(\ve{v}_\Theta^\text{unpruned}(x_t,t))\Vert_2^2], \\
    \end{split}
\end{equation}
where $\texttt{SG}(\cdot)$ denotes the stop gradient operator. This 'soft' supervision serves to enhance the pruned model's robustness. Thus, the final loss can be denoted as
\begin{equation}
    \begin{split}
        & \mathcal{L}^\text{stage-II} = \mathcal{L}^\text{stage-II}_\text{distill} + \mathcal{L}^\text{stage-II}_\text{training}.\\
    \end{split}
\end{equation}
Our ablation experiments show that the model achieves optimal performance when $\alpha=1$. This setting completely removes the ground truth video supervision, relying solely on the distillation loss from the unpruned model. This result demonstrates that during this process, 'soft' supervision from the teacher model output is more effective than 'rigid' supervision from the ground truth video.

\begin{table*}[h]
\centering
\caption{TI2V quantitative results comparison between FastLightGen and other popular large-scale VDMs on general I2V-VBench. Experimental results for the other comparison methods were directly extracted from the official VBench website. FastLightGen outperforms all prior open-source state-of-the-art image-to-video models, including the distilled teacher, Wan-TI2V.}
\label{tab:i2v-comparison}
\vspace{-10pt}
\resizebox{0.85\linewidth}{!}{
\renewcommand{\arraystretch}{0.85}
    \begin{tabular}{lcccccccl} % 'l' for left-aligned Method column, 7 'c' for metrics, final 'l' for average
        \toprule 
        \multirow{1}{*}{Method} & \makecell{i2v\\subject} & \makecell{subject\\consistency} & \makecell{motion\\smoothness} & \makecell{dynamic\\degree} & \makecell{aesthetic\\quality} & \makecell{imaging\\quality} & \multirow{1}{*}{average} \\
        \midrule
        % ---------------------- Data Rows ----------------------
        CogVideoX-I2V-SAT~\citep{yang2024cogvideox} & \best{0.977} & 0.955 & 0.984 & 0.365 & 0.598 & 0.676 & 0.759 \\
        I2Vgen-XL~\citep{2023i2vgenxl}           & 0.975 & 0.964 & 0.983 & 0.250 & 0.653 & 0.699 & 0.753 \\
        SEINE-512x320~\citep{chen2023seine}      & 0.966 & 0.942 & 0.967 & 0.343 & 0.584 & 0.710 & 0.751 \\
        VideoCrafter-I2V~\citep{chen2024videocrafter2}   & 0.912 & \best{0.979} & 0.980 & 0.226 & 0.608 & \best{0.717} & 0.736 \\
        SVD-XT-1.0~\citep{blattmann2023stable}         & 0.975 & 0.955 & 0.981 & \best{0.524} & 0.602 & 0.698 & 0.789 \\
        Step-Video-TI2V~\citep{stepfunvideo}    & 0.955 & 0.960 & \best{0.992} & 0.488 & 0.630 & 0.704 & 0.788 \\
        Wan-TI2V (Teacher)~\citep{wan2025wan}            & 0.961    &0.970  & 0.982    & 0.461    & 0.653   & 0.711    & {0.790}   \\
        FastLightGen       & 0.952 & 0.956 & 0.983 & 0.500 & \best{0.656} & \best{0.717} & \best{0.794} \\
        \bottomrule
    \end{tabular}
}
\vspace{-10pt}
\end{table*}

\begin{table*}[t]
\centering
\caption{Quantitative comparison of FastLightGen against competing acceleration algorithms on WanX-TI2V. For all metrics, higher is better (↑), except for time spent, where lower is better (↓). FastLightGen outperforms DMD2, LCM, F3-Pruning and ICMD in both speed and performance. FastLightGen achieves an empirical speedup of approximately 35$\times$ over the Euler baseline without performance degradation.}
\label{tab:acceleration-comparison}
\vspace{-10pt}
\resizebox{0.85\linewidth}{!}{
\renewcommand{\arraystretch}{0.85}
    \begin{tabular}{lcccccccll} % 'l' for left-aligned Method column, 6 'c' for metrics, final 'l' for time spent
        \toprule 
        \multirow{1}{*}{Method} & \makecell{i2v\\subject} & \makecell{subject\\consistency} & \makecell{motion\\smoothness} & \makecell{dynamic\\degree} & \makecell{aesthetic\\quality} & \makecell{imaging\\quality} & \makecell{average} & \makecell{time\\spent} \\
        \midrule
        % ---------------------- Data Rows ----------------------
        Euler & \best{0.961} & \best{0.970} & 0.982 & 0.461 & 0.653 & 0.711 & 0.790 & 885.3s \\
        DMD2~\citep{dmd2}                & 0.948 & 0.946 & 0.977 & 0.160 & 0.583 & 0.683 & 0.716 & 35.4s \\
    LCM~\citep{icml23_consistency}                & 0.945 & 0.942 & 0.979 & 0.003 & 0.570 & 0.665 & 0.684 & 35.4s \\
MagicDistillation~\citep{shao2025magicdistillation}  & {0.956} & 0.958 & 0.980 & \best{0.561} & 0.634 & 0.701 & \best{0.798} & 35.4s \\
        ICMD~\citep{wu2024individual}               & 0.951 & 0.949 & 0.982 & 0.363 & 0.653 & 0.715 & 0.769 & 396.2s \\
        F3-Pruning~\citep{su2024f3}               & 0.938 & 0.935 & 0.977 & 0.300 & 0.651 & 0.706 & 0.751 & 396.2s \\
        FastLightGen       & 0.952 & {0.956} & \best{0.983} & 0.500 & \best{0.656} & \best{0.717} & 0.794 & \best{28.3s} \\
        \bottomrule
    \end{tabular}
}
\vspace{-10pt}
\end{table*}

\begin{table}[t]
\centering
\caption{User study on models trained with varying distill and training weights. Users rated models on a 5-point scale (higher is better) for action quality and visual quality. The results indicate that the best overall performance is achieved when the distill weight $\alpha$ is set to 1 and the training weight $1-\alpha$ is 0.}
\vspace{-10pt}
\resizebox{1.0\linewidth}{!}{
\renewcommand{\arraystretch}{0.85}
    \begin{tabular}{cccc}
    \toprule
    \makecell{distill weight ($\alpha$)} & \makecell{training weight ($1-\alpha$)} & action score & vision score \\ 
    \hline
    0.0 & 1.0 & 2.86 & 3.41 \\
    0.3 & 0.7 & 2.88 & 3.52 \\
    0.5 & 0.5 & 2.85 & 3.55 \\
    0.7 & 0.3 & 2.89 & 3.58 \\
    1.0 & 0.0 & \best{2.92} & \best{3.61} \\ 
       \bottomrule
    \end{tabular}
}
\label{tab:alpha-distill}
\vspace{-10pt}
\end{table}

\subsection{Stage III: Performing a Fine-grained Distribution Matching}

\begin{figure}[t]
    \centering
    \includegraphics[width=1.0\linewidth,trim={5 0 0cm 0},clip]{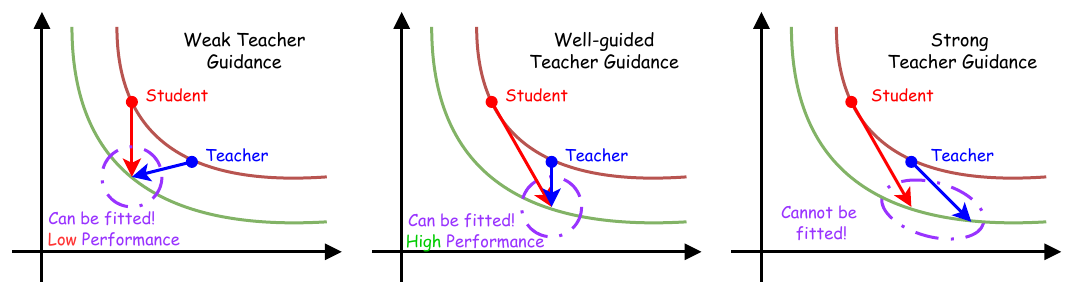}
    \vspace{-20pt}
    \caption{Well-guided teacher guidance {outperforming} approaches that rely solely on weak and strong teachers. Specifically, weak teacher supervision may enable student models to mimic the teacher's outputs, yet this often leads to suboptimal performance. Conversely, overly strong teachers might provide excessively complex supervision signals, hindering effective knowledge transfer and efficient student learning.}
    \label{fig:well_guided_teacher_guidance}
    \vspace{-10pt}
\end{figure}

% Although Stages I and II yield a robust, dynamically pruned model, its outputs suffer from low fidelity and poor diversity, failing to meet practical requirements. Furthermore, a coordinated distillation of sampling steps and model size has not yet been performed, meaning the model cannot produce high-quality videos when restricted to few-step (e.g., 4-step) sampling under a high pruning ratio.

Stage III introduces fine-grained distribution matching for co-distillation of step and size. Within this framework, the pruned model will serve as a `few-step generator', while a hybrid of the unpruned and pruned models will function as the `real DiT'. This approach provides precise and reliable teacher guidance, enabling the coordinated distillation of both steps and parameters. Specifically, we first define the few-step generator, real DiT and fake DiT as $G_\phi$, $\ve{v}_\Theta^\textrm{real}$ and $\ve{v}_\Theta^\textrm{fake}$, respectively. The distribution matching loss is formulated as the expectation over $t$ of an approximate Kullback-Liebler (KL) divergence between the diffused target distribution $p_\text{real}$ and the diffused generator output distribution $p_\text{fake}$. This loss can be computed as the difference of 2 score functions:
\begin{equation}
\fontsize{9pt}{11pt}\selectfont
    \begin{split}
        &\nabla_\phi \mathcal{L}^\text{stage-III}_\textrm{dm} \triangleq \mathbb{E}_{t} \nabla_\phi \mathcal{D}_\textrm{KL}\left(p_\textrm{fake}\Vert p_\textrm{real}\right) \approx \\
        &- \mathbb{E}_t \int_\epsilon\Big([ \sigma_t\ve{v}_\Theta^\textrm{fake}(\ve{x}_t,t)-\sigma_t\ve{v}_\Theta^\textrm{real}(\ve{x}_t,t)]\frac{\partial G_\phi(\epsilon)}{\partial \phi}d\epsilon\Big).\\
    \end{split}
    \label{eq:dmd_loss}
\end{equation}
Here, both the few-step generator and fake DiT are initialized with $\ve{v}_\Theta^\text{pruned}$. The fake DiT is trained with a standard denoising objective to model the output distribution of the few-step generator. During training, the `real DiT' parameters are held frozen, whereas the `fake DiT' parameters are actively updated. The real DiT leverages outputs from both $\ve{v}_\Theta^\text{pruned}$ and $\ve{v}_\Theta^\text{unpruned}$, which can be denoted as
\begin{equation}
\fontsize{9pt}{11pt}\selectfont
    \begin{split}
        & \beta_1^\text{cur} \sim\mathcal{U}[\beta_1-a_1,\beta_1+b_1], \beta_2^\text{cur} \sim\mathcal{U}[\beta_2-a_2,\beta_2+a_2],  \\
        & \ve{v}_\Theta^\textrm{real}(\ve{x}_t,t) = \beta_1^\text{cur}(\ve{v}_\Theta^\textrm{pruned}(\ve{x}_t,t, c) - \ve{v}_\Theta^\textrm{pruned}(\ve{x}_t,t, \emptyset))  \\
        & + \beta_2^\text{cur}(\ve{v}_\Theta^\textrm{unpruned}(\ve{x}_t,t, c) - \ve{v}_\Theta^\textrm{pruned}(\ve{x}_t,t, c)) + \ve{v}_\Theta^\textrm{pruned}(\ve{x}_t,t, \emptyset), \\
    \end{split}
    \label{eq:output_real_dit}
\end{equation}
where $\mathcal{U}[\cdot,\cdot]$, $\beta_1$, $\beta_2$, $c$, $\emptyset$ refer to the uniform distribution, the inter CFG, the intra CFG, the text prompt, the null prompt, respectively. Additionally, $a_1$, $a_2$, $b_1$, and $b_2$ are used to describe the uniform distribution. Specifically, $a_1$, $a_2$, $b_1$, and $b_2$ are empirically set to 1, 0.1, 1, and 0.1, respectively. Sampling CFG values from a uniform distribution serves to improve our method's robustness. We term this mechanism, formulated in Eq.~\ref{eq:output_real_dit}, the well-guided teacher guidance. Intuitively, the term $\beta_2^\text{cur}(\ve{v}_\Theta^\textrm{unpruned}(\ve{x}_t,t, c) - \ve{v}_\Theta^\textrm{pruned}(\ve{x}_t,t, c))$ is analogous to CFG. The coefficient $\beta_2^\text{cur}$ functions as a guidance scale: a larger value steers the resulting output closer to that of the unpruned model $\ve{v}_\Theta^\textrm{unpruned}$, while a smaller value biases the output toward the pruned model $\ve{v}_\Theta^\textrm{pruned}$. As shown in Fig.~\ref{fig:well_guided_teacher_guidance}, our well-guided teacher guidance avoids two common failure modes: supervision from an overly weak teacher, which is ineffective, and guidance from an overly strong teacher, which the student model cannot effectively follow. By tuning the inter CFG and intra CFG, we can independently control two orthogonal aspects of the teacher: the strength of its text-conditional guidance and the influence from the unpruned model. This allows us to find an optimal combination of these two factors, creating a carefully calibrated supervision signal that is ideally suited for the student.

%% file: sec/4_experiment.tex
\section{Experiments}
\label{sec:experiment}

\begin{table*}[t]
\centering
\caption{Quantitative Comparison of FastLightGen under Varying Distill and Training Weights (Teacher: WanX-TI2V). The optimal configuration, with the distill weight set to 1 and the training weight set to 0, enables this setting to outperform all other hyperparameter settings across key metrics, including i2v subject, subject consistency, motion smoothness, dynamic degree, and aesthetic quality.}
\label{tab:distill_ratio_eval}
\vspace{-10pt}
\resizebox{0.85\linewidth}{!}{
\renewcommand{\arraystretch}{0.85}
    \begin{tabular}{cccccccccc} % 分割线在 train weight 和 i2v subject 之间
        \toprule 
        \makecell{distill weight\\($\alpha$)} & \makecell{training weight\\($1-\alpha$)} & \makecell{i2v\\subject} & \makecell{subject\\consistency} & \makecell{motion\\smoothness} & \makecell{dynamic\\degree} & \makecell{aesthetic\\quality} & \makecell{imaging\\quality} & \makecell{temporal\\flickering} & \makecell{average} \\
        \midrule % 表头下方的 \midrule
        
        % Data Rows
        Teacher & N/A & 0.959 & 0.965 & {0.991} & 0.309 & {0.663} & 0.709 & 0.983 & 0.797 \\
        0.0 & 1.0 & 0.957 & 0.964 & 0.990 & 0.205 & 0.648 & {0.711} & {0.984} & 0.780 \\
        0.3 & 0.7 & 0.959 & 0.969 & \best{0.991} & 0.224 & 0.653 & \best{0.712} & \best{0.985} & 0.785\\
        0.5 & 0.5 & 0.959 & 0.968 & \best{0.991} & 0.244 & 0.652 & \best{0.712} & 0.983 & 0.780 \\
        0.7 & 0.3 & 0.959 & 0.965 & 0.990 & 0.256 & 0.654 & \best{0.712} & 0.983 & 0.788 \\
        1.0 & 0.0 & \best{0.961} & \best{0.970} & \best{0.991} & \best{0.268} & \best{0.655} & {0.711} & {0.984} & \best{0.791} \\
        \bottomrule
    \end{tabular}
}
\vspace{-10pt}
\end{table*}

\begin{table*}[t]
\centering
\caption{Quantitative results comparison of $\beta_1$ and $\beta_2$ on WanX-TI2V. Fixing $\beta_2$ at 0.25 yields the best overall performance when $\beta_1$ is set $\beta_1$ to 3.5. While $\beta_2$ values of 0.5, 0.75, and 1.0 achieve higher dynamic degree metrics when $\beta_1$ is fixed at 3.5, the synthesized videos suffer from noticeable jitter and motions that violate physical laws upon visual inspection. Thus, we select $\beta_2=0.25$ as the optimal choice.}
\label{tab:beta12_wanx}
\vspace{-10pt}
\resizebox{0.85\linewidth}{!}{
\renewcommand{\arraystretch}{0.85}
    \begin{tabular}{cccccccccc}
        \toprule 
        \makecell{Inter CFG\\($\beta_1$)} & \makecell{Intra CFG\\($\beta_2$)} & \makecell{i2v\\subject} & \makecell{subject\\consistency} & \makecell{motion\\smoothness} & \makecell{dynamic\\degree} & \makecell{aesthetic\\quality} & \makecell{imaging\\quality} & \makecell{temporal\\flickering} & \makecell{average} \\
        \midrule % 表格中间细线
        
        % Group 1: Fixed \beta_1 = 3.5, varying \beta_2
        \multicolumn{9}{l}{\textbf{Fixed $\beta_1 = 3.5$:}} \\ % 添加分组标签
        \midrule 
        3.5 & 0.00 & 0.952 & 0.956 & 0.983 & 0.459 & 0.646 & 0.710 & \best{0.978} & 0.812 \\
        3.5 & 0.25 & 0.952 & 0.956 & 0.983 & 0.500 & \best{0.656} & \best{0.717} & 0.975 & 0.820 \\
        3.5 & 0.50 & \best{0.957} & \best{0.959} & \best{0.986} & 0.518 & 0.638 & \best{0.717} & 0.976 & 0.822 \\
        3.5 & 0.75 & 0.927 & 0.886 & 0.961 & \best{1.000} & 0.584 & 0.666 & 0.918 & \best{0.848} \\
        3.5 & 1.00 & 0.909 & 0.876 & 0.945 & \best{1.000} & 0.572 & 0.652 & 0.909 & 0.838 \\ 
        \midrule % 分割线
        
        % Group 2: Fixed \beta_2 = 0.25, varying \beta_1
        \multicolumn{9}{l}{\textbf{Fixed $\beta_2 = 0.25$:}} \\ % 添加分组标签
        \midrule
        1.5 & 0.25 & \best{0.957} & \best{0.973} & \best{0.988} & 0.179 & \best{0.663} & 0.711 & \best{0.985} & 0.779 \\
        2.5 & 0.25 & \best{0.957} & 0.966 & \best{0.988} & 0.268 & 0.652 & 0.714 & 0.983 & 0.790 \\
        3.5 & 0.25 & 0.952 & 0.956 & 0.983 & \best{0.500} & 0.656 & \best{0.717} & 0.975 & \best{0.820} \\
        4.5 & 0.25 & \best{0.957} & 0.965 & 0.987 & 0.285 & 0.624 & 0.713 & 0.983 & 0.788 \\ 
        \bottomrule % 表格底部粗线
    \end{tabular}
}
\vspace{-10pt}
\end{table*}

We evaluated all HunyuanVideo-ATI2V experiments on our customized VBench-I2V, while all WanX-TI2V experiments were assessed using VBench-I2V. The 'ATI2V' designation signifies that the model accepts audio, text, and image inputs to generate videos. This model was fine-tuned from the HunyuanVideo-T2V checkpoint on a proprietary dataset, equipping it with audio and image encoding capabilities. In this section, we first describe our experimental setup, including dataset construction and hyperparameter settings. We then present and analyze key comparative results, concluding with an analysis of our ablation studies.

% \paragraph{Our Dataset.} 
In this study, we utilize a dataset derived from two primary sources: open-source collections and user-generated content (UGC). The open-source component consists mainly of landscape-oriented videos from large-scale datasets, including OpenVid-1M~\citep{openvid}, OpenHumanVid~\citep{openhumanvid}, Intern-4K~\citep{intern4k}, and Koala-36M~\citep{koala}. Our UGC data is primarily sourced from YouTube and TikTok; notably, TikTok provides a rich corpus of vertical-aspect-ratio videos, significantly enhancing data diversity. To ensure quality, all data is filtered using the Dover score~\citep{dover}, with videos scoring below 80 being excluded. The ethics statement section can be found in Appendix~\ref{app:ethics_statement}. The benchmarking and hyperparameter settings are specified in the Appendix.~\ref{app:benchmark-hyperparameter}.

% \subsection{Main Comparison}

\noindent{\bf Main Result.} FastLightGen significantly outperforms the state-of-the-art open-source video generation models and other accelerated sampling algorithms, as presented in Tables~\ref{tab:i2v-comparison} and~\ref{tab:acceleration-comparison}. Specifically, Table~\ref{tab:i2v-comparison} contrasts FastLightGen with a suite of models, including CogVideoX-I2V-SAT~\citep{cogvideox}, I2Vgen-XL~\citep{2023i2vgenxl}, SEINE-512x320~\citep{chen2023seine}, VideoCrafter-I2V~\citep{chen2024videocrafter2}, SVD-XT-1.0~\citep{blattmann2023stable}, Step-Video-TI2V~\citep{stepfunvideo}, and our teacher model, WanX-TI2V~\citep{wan2025wan}. FastLightGen surpassed all other models in both aesthetic and imaging quality, underscoring its exceptional visual fidelity. Furthermore, our model even outperformed its own teacher, WanX-TI2V, on the overall average score.

In Table~\ref{tab:acceleration-comparison}, we compare FastLightGen against step distillation algorithms (i.e., DMD2~\citep{dmd2}, LCM~\citep{icml23_consistency}, MagicDistillation~\citep{shao2025magicdistillation}) and the model pruning algorithms (i.e., F3-Purning~\citep{su2024f3} and ICMD~\citep{wu2024individual}). FastLightGen outperforms all competing algorithms in subject consistency, motion smoothness, aesthetic quality, and imaging quality. Furthermore, its average score surpasses all methods except MagicDistillation, trailing by a negligible 0.004 points while reducing inference time by 7.1 seconds. This series of comparative experiments fully demonstrates the generalizability and superiority of FastLightGen.

\noindent{\bf Ablation Study.} In Stage II, we introduce two loss terms: supervision from the ground truth video on $\ve{v}_\Theta^\text{pruned}$, and a distillation loss from $\ve{v}_\Theta^\text{unpruned}$. The parameter $\alpha$ controls the relative weight of these two signals. As illustrated in Tables~\ref{tab:distill_ratio_eval} and~\ref{tab:alpha-distill}, FastLightGen achieves optimal performance at $\alpha\!=\!1$, a setting that removes ground truth supervision entirely and relies solely on distillation. At this value, FastLightGen outperforms other settings on VBench-I2V, achieving the highest overall average score. Furthermore, it surpasses other configurations in human-rated metrics, namely the action and vision scores. We therefore set $\alpha\!=\!1$ as the default in all comparative experiments.

The inter CFG $\beta_1$ and intra CFG $\beta_2$ are critical hyperparameters in Stage III, controlling the strength of text-conditional guidance and the strength of the teacher model (i.e., real DiT), respectively. Intuitively, a larger $\beta_1$ can lead to exaggerated motion and oversaturation. Conversely, a larger $\beta_2$ biases the model's output towards the unpruned model ($\ve{v}_\Theta^\text{unpruned}$) and away from the pruned model ($\ve{v}_\Theta^\text{pruned}$), thereby strengthening the `real DiT' component. We conduct ablation experiments on both HunyuanVideo-ATI2V and WanX-TI2V, with the results presented in Table~\ref{tab:beta12_hunyuan} (see Appendix) and Table~\ref{tab:beta12_wanx}. For $\beta_1$, the optimal values are 2.0 (HunyuanVideo) and 3.5 (WanX). This represents a trade-off: although Table~\ref{tab:beta12_hunyuan} indicates that dynamic degree continues to rise with $\beta_1$, our qualitative assessment found that higher values produce uncontrollable videos with severe artifacts, such as jitter and the hallucination of objects not in the prompt. For $\beta_2$, the effective range was much smaller. We observe an interesting, divergent trend: increasing $\beta_2$ decreases the dynamic degree for HunyuanVideo-ATI2V, whereas it increases it for WanX-TI2V. Despite this difference, $\beta_2=0.25$ substantiates optimal for both. Further analysis of $\beta_2$ can be found in Appendix~\ref{app:beta2_analysis}.

Additional ablation studies on ``The Importance of Stage III'', ``Pruning Ratio'', and ``Teacher Category'' are available in Appendix~\ref{app:ablation-studies}.

\noindent{\bf Visualization.} Fig.~\ref{fig:visualization_best_performance} showcases qualitative results from FastLightGen trained with optimal parameters. The figure demonstrates that our algorithm effectively handles diverse scenarios, including e-commerce, food vlogging, natural landscapes, and dance performances, producing high-fidelity videos with excellent motion quality.

%% file: sec/5_conclusion.tex
\section{Conclusion}

In this paper, we propose FastLightGen, a novel step-and-size co-distillation algorithm for VDMs. Our method comprises three stages: \textit{\textbf{1)}} identifying and pruning unimportant layers, \textit{\textbf{2)}} training a dynamically pruned model, and \textit{\textbf{3)}} distilling the model via fine-grained distribution matching. Additionally, we introduce a well-guided teacher guidance that dynamically adjusts the guidance intensity to match the student model's capacity, thereby enabling the distillation of an optimal student model. For future work, we plan to extend FastLightGen beyond the TI2V (i.e., text-and-image-to-video) task to broader applications like text-to-video and video-to-video, further advancing the development of lightweight video sampling.

\section*{Acknowledgments}
This work was supported by the National Natural Science Foundation of China under Grant No.~62506317.

%% file: sec/appendix.tex
\clearpage
\onecolumn
\appendix

\section{Benchmarks and Hyperparameter Settings}
\label{app:benchmark-hyperparameter}

\paragraph{Our Benchmark.} Our benchmarking strategy is twofold: we use the generic VBench-I2V~\citep{vbench} to evaluate WanX-I2V and a customized VBench-I2V to assess HunyuanVideo-ATI2V. Regarding the generic VBench-I2V, the original VBench \citep{vbench} was designed exclusively for T2V generators and could not assess I2V performance. Consequently, VBench-I2V \citep{vbench++} was developed as an extension, building upon the foundational VBench framework. This enhancement adds over 1118 new text-image pairs, facilitating a comprehensive evaluation of I2V generators across multiple dimensions—such as visual quality, motion dynamics, and subject consistency—with all testing conducted at a 16:9 aspect ratio. Our customized VBench-I2V is specifically designed for testing digital human scenarios. We collected 58 widescreen images featuring real people and anime characters and generated corresponding prompts using InternVL-26B. Similar to the generic VBench-I2V, we employed a 16:9 aspect ratio and repeated each sample five times to mitigate random errors.

\paragraph{Hyperparameter Settings for Training.} All experiments were conducted on 16 NVIDIA H100 GPUs (2 nodes of 8) with a per-GPU batch size of 1, resulting in a global batch size of 16. We employed DeepSpeed ZeRO-3 and AdamW with CPU offloading to mitigate GPU memory overhead. For Stage II, we set the learning rate to $1e\!-\!5$ and trained for 4000 iterations. For Stage III, the learning rate was $5e\!-\!7$ and training ran for 1000 iterations. Specifically, Stage II consumes approximately 64 GPU days, while Stage III requires approximately 16 GPU days. The video memory footprint for both stages remains under 80GB. This shorter duration for Stage III was chosen empirically, as we observed that training beyond this point produced overly abrupt video motions and oversaturated colors. The hyperparameters $(\alpha, \beta_1, \beta_2)$ were set to $(1, 2.0, 0.25)$ for the HunyuanVideo-TI2V experiments and $(1, 3.5, 0.25)$ for the WanX-TI2V experiments.

\paragraph{Hyperparameter Settings for Evaluation.} For the general-purpose VBench-I2V benchmark, we adopted the official evaluation prompts and the standard 16:9 resolution. On our customized VBench-I2V benchmark, identical prompts and a 16:9 resolution were used for all evaluations. To account for generative stochasticity, we generated five distinct videos for each prompt using different random seeds.

\section{Additional Ablation Study}
% \subsection{Ablation Study}
\label{app:ablation-studies}

\begin{table*}[t]
\centering
\caption{Quantitative comparison of Stage III application and varying pruning ratios on WanX-TI2V. Applying Stage III, which performs coordinated distillation of sampling step and model size, results in slight declines across metrics like i2v subject, subject consistency, motion smoothness, and temporal flickering. Crucially, however, it achieves significant improvements in dynamic degree, aesthetic quality, and imaging quality. We also observe that larger pruning ratios correlate with poorer model performance overall.}
\label{tab:vbench2-dmd}
\resizebox{0.9\linewidth}{!}{
    % 注意这里使用了 ccc|ccccccc，所以竖线在第三列和第四列之间
    \begin{tabular}{ccccccccccc}
        \toprule 
        \makecell{Stage\\III} & \makecell{sampling\\steps} & \multicolumn{1}{c}{\makecell{pruning\\ratio}} & \multicolumn{1}{c}{\makecell{i2v\\subject}} & \makecell{subject\\consistency} & \makecell{motion\\smoothness} & \makecell{dynamic\\degree} & \makecell{aesthetic\\quality} & \makecell{imaging\\quality} & \makecell{temporal\\flickering} & \makecell{average} \\
        \midrule % 表头下方的 \midrule 完美绘制
        
        % 30% Pruning Ratio
        \xmark & 28 & 30\% & \best{0.961} & \best{0.970} & \best{0.991} & 0.268 & 0.655 & 0.711 & \best{0.984} & 0.791 \\
        \cmark & 4 & 30\% & 0.960 & 0.966 & 0.987 & \best{0.378} & \best{0.658} & \best{0.712} & 0.977 & \best{0.805} \\
        \midrule % 分割线
        
        % 50% Pruning Ratio
        \xmark & 28 & 50\% & \best{0.958} & \best{0.964} & \best{0.988} & 0.174 & 0.632 & 0.692 & \best{0.989} & 0.771 \\
        \cmark & 4 & 50\% & 0.957 & 0.958 & 0.971 & \best{0.273} & \best{0.649} & \best{0.701} & 0.976 & \best{0.784} \\
        \midrule % 分割线
        
        % 70% Pruning Ratio
        \xmark & 28 & 70\% & \best{0.959} & \best{0.959} & \best{0.965} & 0.064 & 0.552 & 0.591 & \best{0.983} & 0.725 \\
        \cmark & 4 & 70\% & 0.955 & 0.958 & 0.953 & \best{0.132} & \best{0.579} & \best{0.619} & 0.970 & \best{0.738} \\
        \bottomrule
    \end{tabular}
}
\end{table*}

\paragraph{The Importance of Stage III.} Stage III is critical, as it performs the truly coordinated distillation of both sampling steps and model size. As demonstrated in Table~\ref{tab:vbench2-dmd}, without this stage, FastLightGen remains an underperforming pruned model; its 28-step sampling performance is markedly inferior to that of the final 4-step distilled model.

\begin{table*}[t]
\centering
\caption{Quantitative results for different pruning ratios on HunyuanVideo-ATI2V. As the pruning ratio increases, we observe a significant degradation in model performance, specifically in dynamic degree, aesthetic quality, and imaging quality.}
\label{tab:pruning_ratio_2}
\resizebox{0.9\linewidth}{!}{
    \begin{tabular}{cccccccccc}
        \toprule
        \makecell{Pruning Ratio} & \makecell{i2v\\subject} & \makecell{subject\\consistency} & \makecell{motion\\smoothness} & \makecell{dynamic\\degree} & \makecell{aesthetic\\quality} & \makecell{imaging\\quality} & \makecell{temporal\\flickering} & \makecell{average} \\
        \midrule
        0\%  & 0.965 & \best{0.984} & 0.993 & \best{0.696} & \best{0.603} & \best{0.722} & \best{0.982} & \best{0.849} \\
        30\% & \best{0.966} & 0.972 & \best{0.996} & 0.622 & 0.571 & 0.693 & \best{0.982} & 0.829 \\
        50\% & 0.948 & 0.979 & \best{0.996} & 0.501 & 0.538 & 0.553 & 0.977 & 0.785 \\
        70\% & 0.949 & 0.967 & 0.993 & 0.249 & 0.465 & 0.519 & 0.952 & 0.728 \\
        \bottomrule
    \end{tabular}
}
\end{table*}

\paragraph{Pruning Ratio.} As shown in Fig.~\ref{fig:distilling_laws}, FastLightGen achieves an optimal balance between acceleration and performance at a 70\% parameter retention rate with 4 distillation steps. The contour plot indicates that performance metrics decline sharply when retention falls below this 70\% threshold, as shown by the denser contour lines. This finding is corroborated by the results in Tables~\ref{tab:vbench2-dmd} and~\ref{tab:pruning_ratio_2}. For instance, Table~\ref{tab:vbench2-dmd} demonstrates that pruning beyond this 30\% limit (i.e., < 70\% retention) causes a severe degradation in dynamic degree, aesthetic quality, and overall fidelity, leading to blurry and uncontrollable outputs.

\begin{table*}[t]
\centering
\caption{Quantitative results comparison of $\beta_1$ and $\beta_2$ on HunyuanVideo-ATI2V. Fixing $\beta_2$ at 0.25 yields the best overall performance when $\beta_1$ is set $\beta_1$ to 2.0. With $\beta_1$ fixed at 2.0, we found that higher values for $\beta_2$ (e.g., 0.5,0.75,1.0) improved aesthetic and image quality but significantly degraded lipsyncing. Therefore, we select $\beta_2$=0.25 as it provides the best trade-off between these two objectives.}
\label{tab:beta12_hunyuan}
\resizebox{0.95\linewidth}{!}{
    \begin{tabular}{ccccccccccc}
        \toprule 
        \makecell{Inter CFG\\($\beta_1$)} & \makecell{Intra CFG\\($\beta_2$)} & \makecell{i2v\\subject} & \makecell{subject\\consistency} & \makecell{motion\\smoothness} & \makecell{dynamic\\degree} & \makecell{aesthetic\\quality} & \makecell{imaging\\quality} & \makecell{temporal\\flickering} & \makecell{average} & \makecell{lipsyncing} \\
        \midrule % 表格中间细线
        
        % Group 1: Fixed \beta_1 = 3.5, varying \beta_2
        \multicolumn{9}{l}{\textbf{Fixed $\beta_1 = 2.0$:}} \\ % 添加分组标签
        \midrule
        2.0  & 0.00  & 0.947 & 0.941 & 0.992 & \best{0.667} & 0.585 & 0.687 & 0.986 & \best{0.829} & 1.642 \\
        2.0  & 0.25  & 0.962 & 0.967 & 0.994 & 0.512 & 0.586 & 0.702 & 0.991 & 0.816 & \best{1.967} \\
        2.0  & 0.50  & 0.962 & 0.962 & 0.994 & 0.217 & 0.584 & 0.693 & 0.990 & 0.771 & 1.702 \\
        2.0  & 1.00  & \best{0.966} & \best{0.970} & \best{0.996} & 0.000 & \best{0.587} & \best{0.704} & \best{0.995} & 0.745 & 1.670 \\
        \midrule % 分割线
        
        % Group 2: Fixed \beta_2 = 0.25, varying \beta_1
        \multicolumn{9}{l}{\textbf{Fixed $\beta_2 = 0.25$:}} \\ % 添加分组标签
        \midrule
        1.0  & 0.25  & \best{0.997} & 0.959 & \best{0.995} & 0.052 & 0.585 & 0.695 & \best{0.991} & 0.753 & 1.851 \\
        2.0  & 0.25  & 0.962 & \best{0.967} & 0.994 & 0.512 & \best{0.586} & \best{0.702} & \best{0.991} & 0.816 & \best{1.967} \\
        3.0  & 0.25  & 0.942 & 0.950 & 0.992 & 0.744 & 0.573 & 0.699 & 0.987 & 0.841 & 1.551 \\
        4.0  & 0.25  & 0.963 & 0.962 & 0.994 & 0.755 & 0.570 & 0.688 & \best{0.991} & \best{0.846} & 1.496 \\
        5.0  & 0.25  & 0.924 & 0.931 & 0.992 & \best{0.837} & 0.561 & 0.678 & 0.986 & 0.844 & 1.250 \\
        \bottomrule
    \end{tabular}
}
\end{table*}

\begin{table*}[t]
\centering
\caption{Quantitative results on HunyuanVideo-ATI2V. This ablation investigates the choice of teacher model for the real DiT and the fake DiT in Stage III. We find that employing a pruned model as the teacher significantly outperforms using the unpruned version. This result suggests that the optimal teacher is not necessarily the most powerful, but rather one whose capacity is better matched to the student.}
\label{tab:teacher_if_should_stronger}
\resizebox{0.9\linewidth}{!}{
    \begin{tabular}{lcccccccc}
        \toprule
        \makecell{Teacher Setting} & \makecell{i2v\\subject} & \makecell{subject\\consistency} & \makecell{motion\\smoothness} & \makecell{dynamic\\degree} & \makecell{aesthetic\\quality} & \makecell{imaging\\quality} & \makecell{temporal\\flickering} & \makecell{average} \\
        \midrule
        \textbf{Pruned}  & 0.947 & \best{0.941} & \best{0.992} & {0.667} & \best{0.585} & \best{0.687} & \best{0.986} & \best{0.829}\\
        \textbf{Unpruned} & \best{0.956} & 0.932 & 0.989 & \best{0.670} & 0.525 & 0.612 & 0.980 & 0.809 \\
        \bottomrule
    \end{tabular}
}
\end{table*}

\paragraph{Teacher Category.} In Table~\ref{tab:teacher_if_should_stronger}, we provide further empirical evidence supporting the assertion that ``stronger teachers do not necessarily benefit students more'' On HunyuanVideo-ATI2V, we compare initializing both components with the unpruned model (i.e., a `strong teacher') versus initializing both with the pruned model (i.e., a `weak teacher'). The results show that the `weak teacher' setup (using pruned models) outperforms the ·strong teacher' setup (using unpruned models) on the overall average score in our customized VBench-I2V evaluation. Therefore, our proposed well-guided teacher guidance is effective precisely because it constructs a teacher model that is optimally suited to the student.

\section{Analysis of $\beta_2$}
\label{app:beta2_analysis}
\begin{figure*}[t!]
    \centering
    \includegraphics[width=1\linewidth]{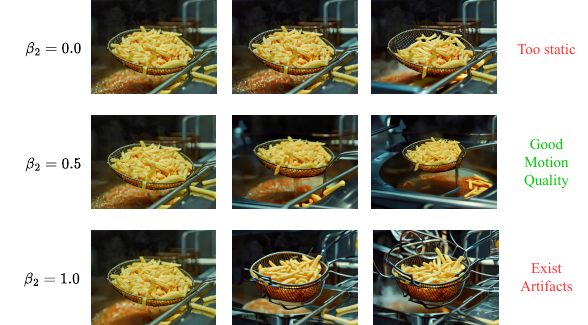}
    \caption{In WanX-TI2V, the $\beta_2$ hyperparameter modulates the dynamic quality of the generated video. A value of $\beta_2=0$ yields overly static results, whereas $\beta_2=1$ introduces abrupt temporal discontinuities and excessive visual artifacts.}
    \label{fig:beta2_analysis}
\end{figure*}

In Stage III, the hyperparameter $\beta_2$ controls the teacher model's interpolation between its unpruned and pruned states. Increasing $\beta_2$ biases the teacher toward the unpruned model, while decreasing $\beta_2$ biases it toward the pruned model. However, a `stronger' teacher (i.e., one closer to the unpruned model) does not necessarily yield superior distillation results, as illustrated in Fig.~\ref{fig:beta2_analysis}. Specifically, setting $\beta_2=0$ (using the fully pruned model) results in overly static videos with poor motion quality. Conversely, setting $\beta_2=1$ (using the fully unpruned model) produces videos with drastic changes and uncontrollable content. Therefore, an appropriate intermediate value for $\beta_2$ is required to construct an optimal teacher for student model distillation.

\section{Ethics Statement}
\label{app:ethics_statement}

We propose FastLightGen, a method for enabling coordinated step and size optimization in large-scale VDMs. The publicly available data used for this method is sourced from online animation clips and authorized public portrait videos. All UGC data is filtered using an automated model to ensure compliance with safety standards and ethical guidelines. We emphasize the responsible deployment and ethical regulation of this technology, aiming to maximize social benefits while proactively monitoring and mitigating potential risks.

\section{Additional Visualization}

\begin{figure*}[t!]
    \centering
    \includegraphics[width=1\linewidth]{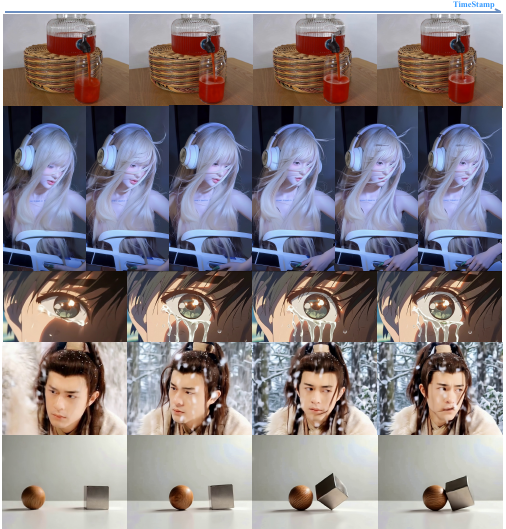}
    \caption{Additional visualization of FastLightGen (i.e., 4-step generator that retains 70\% of the parameters).}
    \label{fig:visualization_pruning_apd1}
\end{figure*}

\begin{figure*}[t!]
    \centering
    \includegraphics[width=1\linewidth]{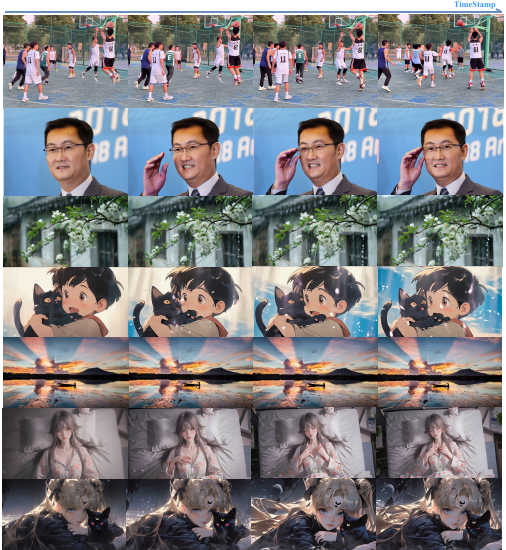}
    \caption{Additional visualization of FastLightGen (i.e., 4-step generator that retains 70\% of the parameters).}
    \label{fig:visualization_pruning_apd2}
\end{figure*}

\begin{figure*}[t!]
    \centering
    \includegraphics[width=1\linewidth]{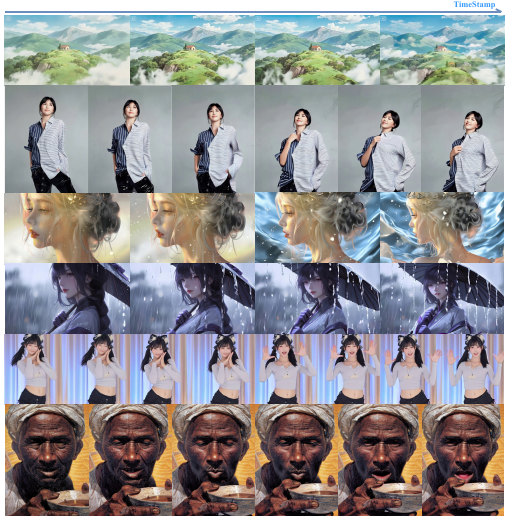}
    \caption{Additional visualization of FastLightGen (i.e., 4-step generator that retains 70\% of the parameters).}
    \label{fig:visualization_pruning_apd3}
\end{figure*}
Here, we present further visualizations of FastLightGen trained with optimal parameters, as shown in Figs.~\ref{fig:visualization_pruning_apd1},~\ref{fig:visualization_pruning_apd2} and~\ref{fig:visualization_pruning_apd3}.